\documentclass[lettersize,journal]{IEEEtran}

\usepackage{amsmath,amsfonts}
\usepackage{algorithmic}
\usepackage{algorithm}
\usepackage{array}
\usepackage[caption=false,font=normalsize,labelfont=sf,textfont=sf]{subfig}
\usepackage{textcomp}
\usepackage{stfloats}
\usepackage{url}
\usepackage{verbatim}
\usepackage{graphicx}
\usepackage{cite}
\usepackage[utf8]{inputenc} 
\usepackage[T1]{fontenc}    
\usepackage{booktabs}       
\usepackage{nicefrac}       
\usepackage{microtype}      
\usepackage{amssymb} 
\usepackage{multirow}
\usepackage{newfloat}
\usepackage{listings}
\usepackage{siunitx} 
\usepackage{colortbl}       
\usepackage{xcolor}         
\usepackage{float}
\usepackage{etoolbox}
\usepackage[colorlinks=true, allcolors=blue]{hyperref}

\graphicspath{ {./images/} }
\begin{document}

\title{FedSDAF: Leveraging Source Domain Awareness for Enhanced Federated Domain Generalization}

\author{
    Hongze Li\textsuperscript{*}, Zesheng Zhou\textsuperscript{*}, Zhenbiao Cao\textsuperscript{*}, Xinhui Li, Wei Chen, and Xiaojin Zhang\textsuperscript{\textdagger}
    \thanks{*The authors contributed equally to this research.}
    \thanks{\textdagger Corresponding author.}
    \thanks{Hongze Li is with the School of Computer Science and Technology, Huazhong University of Science and Technology, Wuhan, Hubei, China (e-mail: lhz13012608937@outlook.com).}
    \thanks{Zesheng Zhou is with the School of Computer Science and Technology, Huazhong University of Science and Technology, Wuhan, Hubei, China (e-mail: zhouzs@tiangong.edu.cn).}
    \thanks{Zhenbiao Cao is with the School of Software Engineering, Huazhong University of Science and Technology, Wuhan, Hubei, China (e-mail: m202477011@hust.edu.cn).}
    \thanks{Xinhui Li is with the School of Computer Science and Technology, Tiangong University, Tianjin, China (e-mail: lixinhui@tiangong.edu.cn).}
    \thanks{Wei Chen is with the School of Software Engineering, Huazhong University of Science and Technology, Wuhan, Hubei, China (e-mail: lemuria\_chen@hust.edu.cn).}
    \thanks{Xiaojin Zhang is with the School of Computer Science and Technology, Huazhong University of Science and Technology, Wuhan, Hubei, China (e-mail: xiaojinzhang@hust.edu.cn).}
}

\markboth{IEEE Transactions on Multimedia, ~Vol.~XX, No.~X, OCTOBER~2025}%
{Li \MakeLowercase{\textit{et al.}}: FedSDAF: Leveraging Source Domain Awareness for Enhanced Federated Domain Generalization}

\maketitle

\begin{abstract}
Traditional Federated Domain Generalization (FedDG) methods focus on learning domain-invariant features or adapting to unseen target domains, often overlooking the unique knowledge embedded within the source domain, especially in strictly isolated federated learning environments. Through experimentation, we discovered a counterintuitive phenomenon: features learned from a complete source domain have superior generalization capabilities compared to those learned directly from the target domain. This insight leads us to propose the Federated Source Domain Awareness Framework (FedSDAF), the first systematic approach to enhance FedDG by leveraging source domain-aware features. FedSDAF employs a dual-adapter architecture that decouples "local expertise" from "global generalization consensus." A Domain-Aware Adapter, retained locally, extracts and protects the unique discriminative knowledge of each source domain, while a Domain-Invariant Adapter, shared across clients, builds a robust global consensus. To enable knowledge exchange, we introduce a Bidirectional Knowledge Distillation mechanism that facilitates efficient dialogue between the adapters. Extensive experiments on four benchmark datasets (OfficeHome, PACS, VLCS, and DomainNet) show that FedSDAF significantly outperforms existing FedDG methods. The source code is available at \url{https://github.com/pizzareapers/FedSDAF}.
\end{abstract}


\section{Introduction}

\begin{figure}[htbp]
  \centering
  \includegraphics[width=\columnwidth]{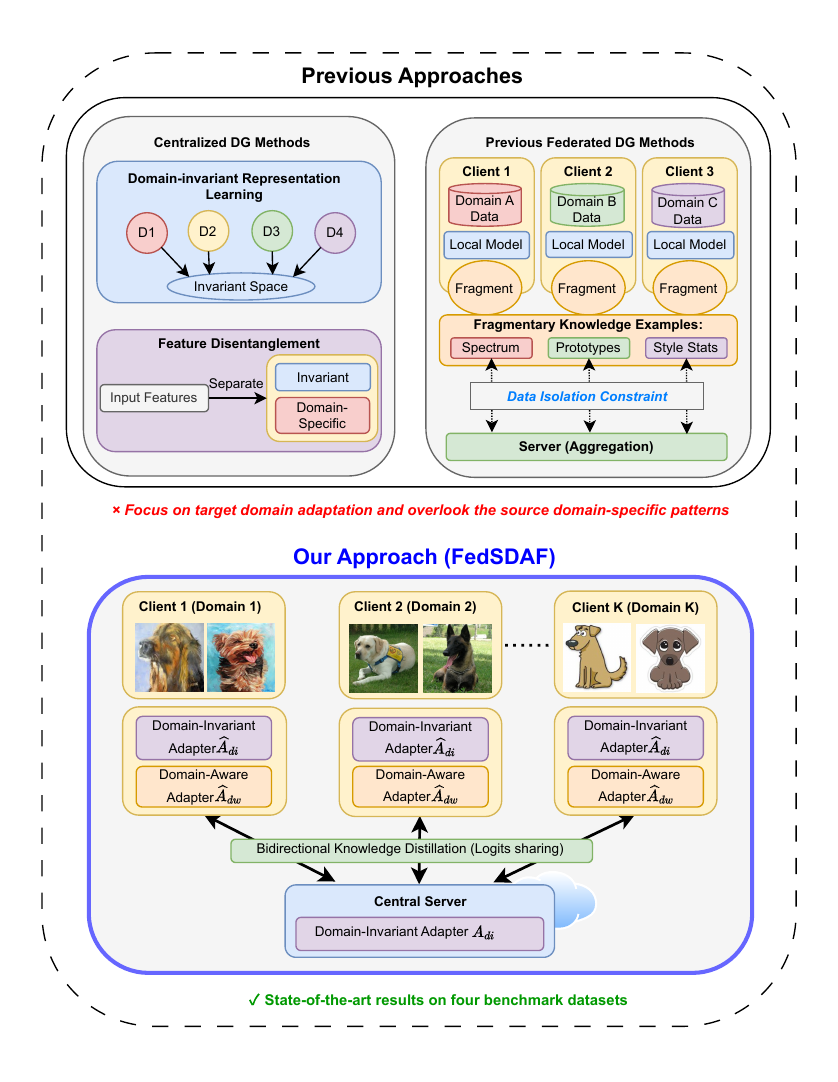}
  \caption{Illustration of domain generalization.}
  \label{fig:head_picture}
\end{figure}

\IEEEPARstart{T}{he} synergy between federated learning (FL) and large-scale pre-trained models \cite{houlsby2019parameterefficienttransferlearningnlp,su2024federatedadaptiveprompttuning} has ushered in a new era of distributed learning paradigms for privacy-sensitive applications. While foundational architectures, particularly Transformers, demonstrate remarkable cross-task adaptability \cite{mahla2025exploringgradientsubspacesaddressing,chen2022fedtunedeepdiveefficient} through parameter-efficient techniques, conventional FL frameworks \cite{McMahan2016CommunicationEfficientLO,10.5555/3495724.3496362} encounter a fundamental limitation: their underlying independent and identically distributed (i.i.d.) assumption stands in stark contrast to the inherent multi-source domain heterogeneity in real-world scenarios \cite{Sahu2018FederatedOI,Peng2020Federated,9577482}. This misalignment presents a dual challenge: not only do client-specific data emerge from distinct domain distributions, but models must also simultaneously address inter-client domain shifts while maintaining robust generalization capabilities for unseen target domains during deployment. These intertwined challenges constitute the core problem in Federated Domain Generalization (FedDG).

Standard domain generalization (DG) methods have advanced through domain-invariant representation learning \cite{10658240,10655267} and feature disentanglement \cite{10655401}. Both \cite{feng-etal-2024-two} and \cite{Li2024DAAdaLD} have adopted the feature disentanglement approach, which separates features into domain-invariant and domain-specific components, utilizing domain-aware features from the target domain to enhance the model's generalization ability in text classification and target detection tasks. However, these studies primarily focus on leveraging domain-aware features in the target domain, without fully exploring the potential value of domain-aware features in the source domain.

The core challenge in Federated Domain Generalization (FedDG) is the strict data isolation among clients, which impedes the model's ability to align knowledge from diverse source domains \cite{Liu2021FedDGFD}. While existing methods attempt to circumvent this by exchanging fragmentary information like spectral data, class prototypes, or style statistics \cite{Liu2021FedDGFD, 10203389, 10030422}, these incomplete representations of knowledge lead to suboptimal fusion during model aggregation.

However, our motivational study reveals a counter-intuitive finding that challenges the conventional focus on domain invariance: in isolated settings, source domain-aware features possess superior generalization capabilities. For instance, on the PACS dataset, an adapter trained solely on the "art painting" source domain achieved 97.07\% accuracy on the unseen "sketch" target, significantly outperforming an adapter trained and tested on "sketch" itself (94.94\%) \cite{8237853, 6751316}. The underlying logic is that a source-aware model, forced to bridge a significant domain gap, must discard superficial source styles (e.g., brushstrokes) and instead learn abstract, essential semantic features, which inherently generalize better to any unseen domain.

This insight reframes the problem: the true challenge is not to eliminate domain differences but to systematically leverage the unique knowledge from each source domain. This creates a core contradiction between preserving client-specific "local expertise" and building a unified "global consensus," a task where simplistic aggregation methods like FedAvg inevitably fail by diluting critical local knowledge.

To resolve this tension, we propose the Federated Source Domain-Aware Framework (FedSDAF). The innovative core of FedSDAF is a unique dual-adapter architecture that transforms this contradiction into a symbiotic relationship through principled decoupling and collaboration. We introduce a fully private, locally-retained Domain-Aware Adapter to act as a "local expert," tasked with mining and safeguarding unique source domain knowledge. In parallel, a shared Domain-Invariant Adapter serves as a "global generalizer," building a robust consensus across all clients. These two roles are interconnected by a novel Bidirectional Knowledge Distillation (BKD) mechanism. This BKD process enables a synergistic dialogue: the local adapter enhances the global model's generalization via its distilled knowledge, while the global model in turn guides the local adapter to prevent overfitting. It is this dynamic interplay between local expertise and global consensus that allows FedSDAF to systematically transform source-specific knowledge from overlooked "noise" into a core "signal" that fundamentally enhances generalization under data isolation.

In this way, FedSDAF transforms source domain knowledge from the "noise" traditionally overlooked by existing methods into a core "signal" that enhances generalization capabilities, systematically addressing the challenges posed by data isolation. Our contributions can be summarized as follows:
\begin{itemize}
    \item \textbf{Proposing a novel FedDG paradigm:} We reveal and validate through rigorous experiments that, in a federated environment with data isolation, source domain-aware features possess greater generalization potential than target domain features. Based on this core insight, we introduce FedSDAF, which systematically utilizes source domain knowledge to enhance model generalization by principled decoupling and collaboration of local domain expertise and global generalization consensus.
    
    \item \textbf{Designing an innovative collaborative learning architecture:} Our dual-adapter and bidirectional knowledge distillation mechanism constructs a dynamic collaborative system. In this system, a private "local expert" adapter and a shared "global generalizer" adapter can mutually promote and co-evolve through a knowledge dialogue mechanism, effectively integrating rich local insights into the global model.
    
    \item \textbf{Achieving state-of-the-art performance on multiple benchmarks:} We conducted extensive experiments on four challenging domain generalization benchmark datasets. The results demonstrate that by systematically harnessing source domain awareness, FedSDAF significantly outperforms existing methods, achieving new state-of-the-art performance and robustly validating the effectiveness and superiority of our proposed approach.
\end{itemize}

\section{Related Work}

\subsection{Federated Learning and Domain Generalization}
Federated Learning (FL) enables collaborative model training on decentralized data without sharing raw information \cite{Tang2023FPPFLFP, Fu2024FedCAFEFC, 10577749, Cai2024TowardsEF}. A primary challenge in FL is the non-Independent and Identically Distributed (non-IID) nature of client data, which impairs model performance \cite{Islam2024FedClustOF, feng2024federated}. While existing methods address this via personalization \cite{Zhang2022FedALAAL} or robust aggregation \cite{NEURIPS2024_29021b06, Li2024MaskedRN}, they often struggle to generalize to entirely unseen distributions \cite{Yuan2021WhatDW, Jiang2024HeterogeneityAwareFD}, a core problem addressed by Domain Generalization (DG).

Standard DG learns robust representations from multiple source domains, often via Domain-Invariant Representation Learning (DIRL) \cite{Xu2022DIRLDR, 9879527, Liu2024FedBCGDCA, 10.1609/aaai.v38i13.29431} or Feature Disentanglement \cite{Gholami2023LatentFD}. These approaches, however, predominantly focus on leveraging features from the \textit{target} domain \cite{feng-etal-2024-two, Li2024DAAdaLD}, leaving the potential of \textit{source-domain-aware} features largely unexplored, especially in federated settings \cite{Zhang2023AggregationOD}.

\subsection{Federated Domain Generalization}
Federated Domain Generalization (FedDG) aims to enhance model generalization to unseen target domains under the data isolation constraints of FL \cite{Zhang2023FederatedDG, Radwan2024FedPartWholeFD}. Prior FedDG studies have explored sharing frequency-space information \cite{Liu2021FedDGFD}, class prototypes \cite{Wan2024FederatedGL}, pixel-level channel statistics \cite{10030422}, or using global consistency augmentation \cite{Liu2024FedGCAGC}. While useful, these methods often exchange shallow feature components (e.g., spectral signatures, style parameters), failing to capture the deeper semantic correlations necessary for robust generalization. Our work addresses this gap by systematically exploiting source domain-aware features.

\section{Methodology}

\subsection{Problem Setup}

In the \textbf{Federated Domain Generalization (FedDG)} setting, we consider a scenario involving \( K \) independent clients. Each client, \( C_k \), exclusively holds a local source dataset \( D_k = \{(x_{ik}, y_{ik})\}_{i=1}^{n_k} \) that is drawn from a unique probability distribution \( P_k(X,Y) \), such that \( P_k \neq P_j \) for any \( k \neq j \). The foundational constraint of this setting is \textbf{Data Isolation}: raw data sharing among clients is strictly prohibited due to privacy concerns. Clients can only collaborate to train a model by exchanging non-sensitive information, such as model parameters or updates, via a central server.
The primary objective in FedDG is to leverage the diverse, isolated source domains \( \{D_k\}_{k=1}^K \) to learn a single, robust global model \( f_{\theta_g} \) that performs well on an entirely unseen target domain \( D_t \). This target domain is drawn from a different distribution \( P_t \), where \( P_t \neq P_k \) for all source clients \( k \in \{1, \ldots, K\} \). The global model \( f_{\theta_g} \) is produced by aggregating the local models trained on each client's private data. This constrained optimization problem is formally expressed as minimizing the expected risk on the target domain:
\[
\theta_g^* = \min_{\theta_g} \mathbb{E}_{(x,y) \sim P_t} \left[ L(f_{\theta_g}(x), y) \right]
\]
where \( L \) represents a predefined loss function.

\subsection{Motivational Study: The Case for Source-Domain Awareness}
The central hypothesis of our work is that in data-isolated settings, features learned from a complete source domain offer a more generalizable foundation than features learned from the target domain itself. To validate this, we conduct a carefully designed motivational study.

\noindent\textbf{Experimental Setup.} We employ a strict leave-one-domain-out protocol on the PACS dataset. For each pair of a single source domain $\mathcal{D}_S$ and a single target domain $\mathcal{D}_T$ (where $\mathcal{D}_S \neq \mathcal{D}_T$), we conduct two parallel and fair experiments:
\begin{enumerate}
    \item \textbf{Source-Aware Generalization:} An adapter is trained \textit{only} on the training set of the source domain $\mathcal{D}_S$ (e.g., Art Painting) and is evaluated on the test set of the unseen target domain $\mathcal{D}_T$ (e.g., Sketch). This measures the ability to generalize knowledge across a significant domain shift.
    \item \textbf{Target-Aware In-Domain Generalization:} A separate adapter is trained \textit{only} on the training set of the target domain $\mathcal{D}_T$ (e.g., Sketch) and is evaluated on its own test set. This represents the standard in-domain performance baseline.
\end{enumerate}
This one-to-one setup ensures that both adapters are trained on datasets of comparable size and diversity, isolating the effect of the learning paradigm itself.

\begin{table}[htbp]
\caption{Performance (\%) of adapters trained on a single source domain (rows) and evaluated on a single target domain (columns).}
\label{tab:motivation}
\centering
\begin{tabular}{l|cccc}
\toprule
\textbf{Adapter Trained on} & \multicolumn{4}{c}{\textbf{Target Domain (Y)}} \\
\textbf{(X)} & Photo & Art & Cartoon & Sketch \\
\midrule
Photo & 96.55 & \textbf{98.92} & \textbf{98.14} & 94.21 \\
Art Painting & \textbf{98.88} & 98.11 & \textbf{98.25} & \textbf{97.07} \\
Cartoon & \textbf{98.03} & \textbf{98.67} & 97.23 & \textbf{95.33} \\
Sketch & \textbf{97.51} & \textbf{97.93} & 94.98 & 94.94 \\
\bottomrule
\end{tabular}
\end{table}

\noindent\textbf{Results and Principled Interpretation.} The results, presented in Table \ref{tab:motivation}, reveal a clear and theoretically grounded trend. In almost all cases, the source-aware adapter (off-diagonal values) outperforms the target-aware adapter (diagonal values) when evaluated on the target domain. For instance, when targeting the Sketch domain, the adapter trained on Art Painting (97.07\%) significantly surpasses the one trained on Sketch itself (94.94\%).

This seemingly counter-intuitive outcome stems from a fundamental learning principle. The target-aware adapter is prone to learning spurious correlations by associating domain-specific styles (e.g., pencil texture in Sketch) with class labels, leading to a brittle representation that overfits. In contrast, the source-aware adapter, forced to generalize across a wide domain gap, cannot rely on the source style (e.g., brushstrokes). This constraint compels the optimization to discard superficial features and instead learn more abstract, semantically invariant representations that capture the essence of the classes. This robust semantic foundation, stripped of domain-specific biases, naturally exhibits superior generalization.

\subsection{Proposed Method}
In this section, we formally introduce our proposed Federated Source Domain-Aware Framework (FedSDAF). The approach comprises two key components: a Knowledge Integrator Adapter (KIA) and a Bidirectional Knowledge Distillation (BKD) process.

\begin{figure}[htbp]
  \centering
  \includegraphics[width=\columnwidth]{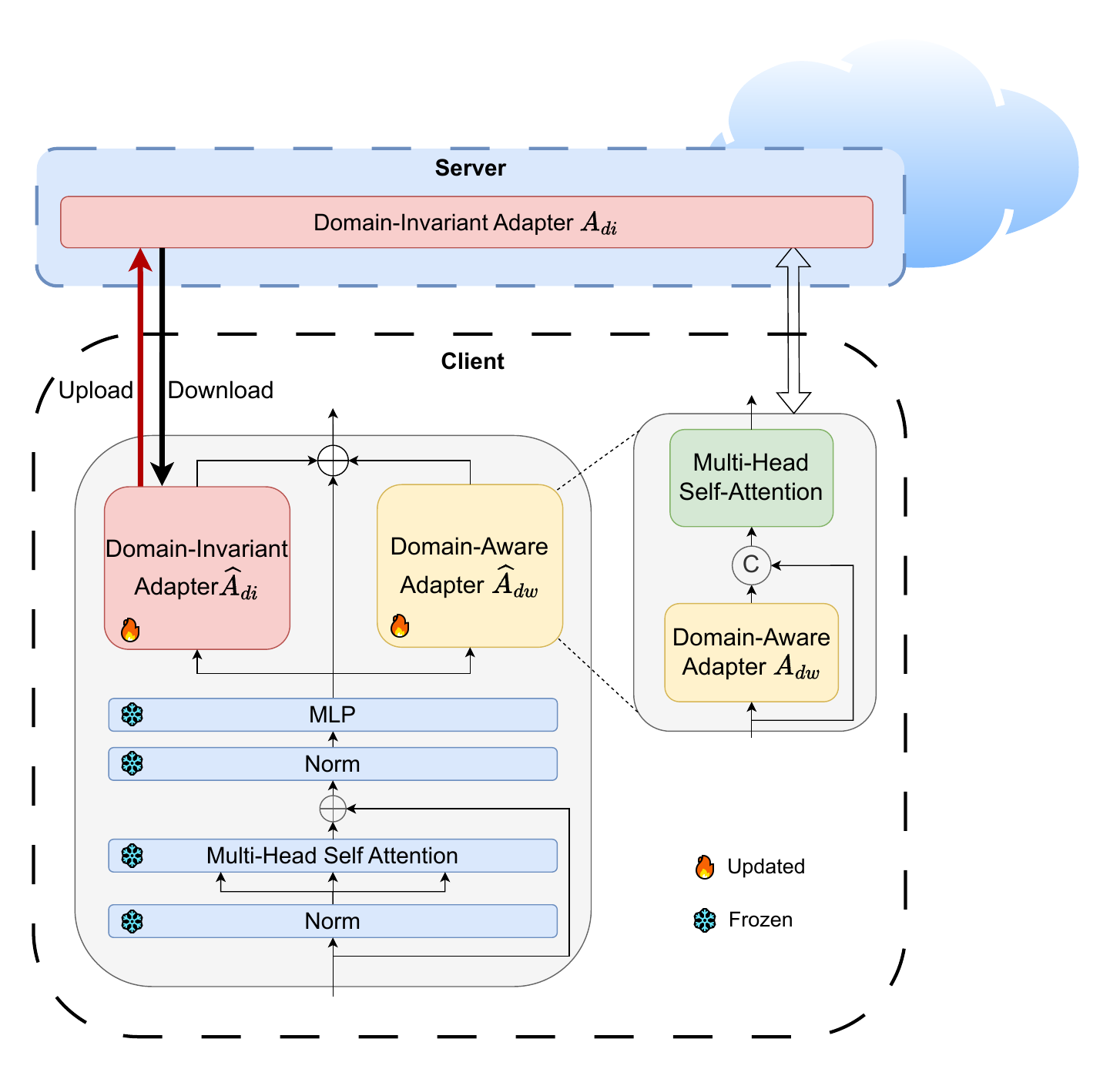}
  \caption{FedSDAF architecture design.}
  \label{fig:architecture}
\end{figure}

\subsubsection{Knowledge Integrator Adapter (KIA)}
The core idea of KIA is to explicitly disentangle and leverage two complementary representation components: domain-invariant features shared by all clients and domain-specific features unique to each client. To achieve this, we introduce two types of adapters into a shared backbone neural model: a domain-invariant adapter $A_{di}$ and a domain-aware adapter $A_{dw}$.

\textbf{Domain-Invariant Adapter $A_{di}$:} The $A_{di}$ aims to capture stable and generalizable representations across all source domains. At each federated communication round $r$, the server maintains a global domain-invariant adapter $A_{di}^r$, which is distributed to all clients for local fine-tuning. Each client $k$ optimizes a local copy of this adapter, denoted as $\hat{A}_{di}^{k,r}$, on its private dataset. After local training, the server aggregates these adapters into an updated global adapter using a weighted averaging scheme, which is essential for maintaining a robust representation that can effectively generalize across various domains.

\textbf{Domain-Aware Adapter $A_{dw}$:} Before the first round of communication, each client initializes a  $A_{dw}$, locally. Unlike the domain-invariant adapter, this adapter is updated strictly locally throughout the training process to force it to capture the domain-aware features of each domain. Its role is to capture subtle, client-specific knowledge from the local domain, thus preserving rich domain-specific information. The domain-aware adapter is crucial for enhancing the model's ability to adapt to the specific characteristics of the client's data.

The KIA, which integrates both $A_{di}$ and $A_{dw}$ (as illustrated in Fig. \ref{fig:architecture}), is inserted following the Feedforward Neural Networks (FFNs) layer of the neural network. To effectively integrate these two sets of features, we employ a Multi-Head Self-Attention (MHSA) mechanism. Let $h$ denote the intermediate representation output from the backbone model. The integrated feature representation on client $k$ is computed as follows:
\begin{equation}
\hat{A}_{dw}(h) = \text{MHSA}(\text{LN}[A_{dw}(h); h]) 
\end{equation}

\begin{equation}
h' = h + \frac{1}{2} \hat{A}_{dw}(h) + \frac{1}{2} \hat{A}_{di}(h)
\end{equation}

This integration $h'$ ensures that in each round of communication, the model effectively assimilates knowledge from the source domain while leveraging the rich contextual information inherent in the domain-aware features. By utilizing MHSA, the model can dynamically weigh the importance of various features from both the source domain and the base model's outputs, facilitating the formation of more adaptive domain-aware features.

\begin{figure}[htbp]
  \centering
  \includegraphics[width=\columnwidth]{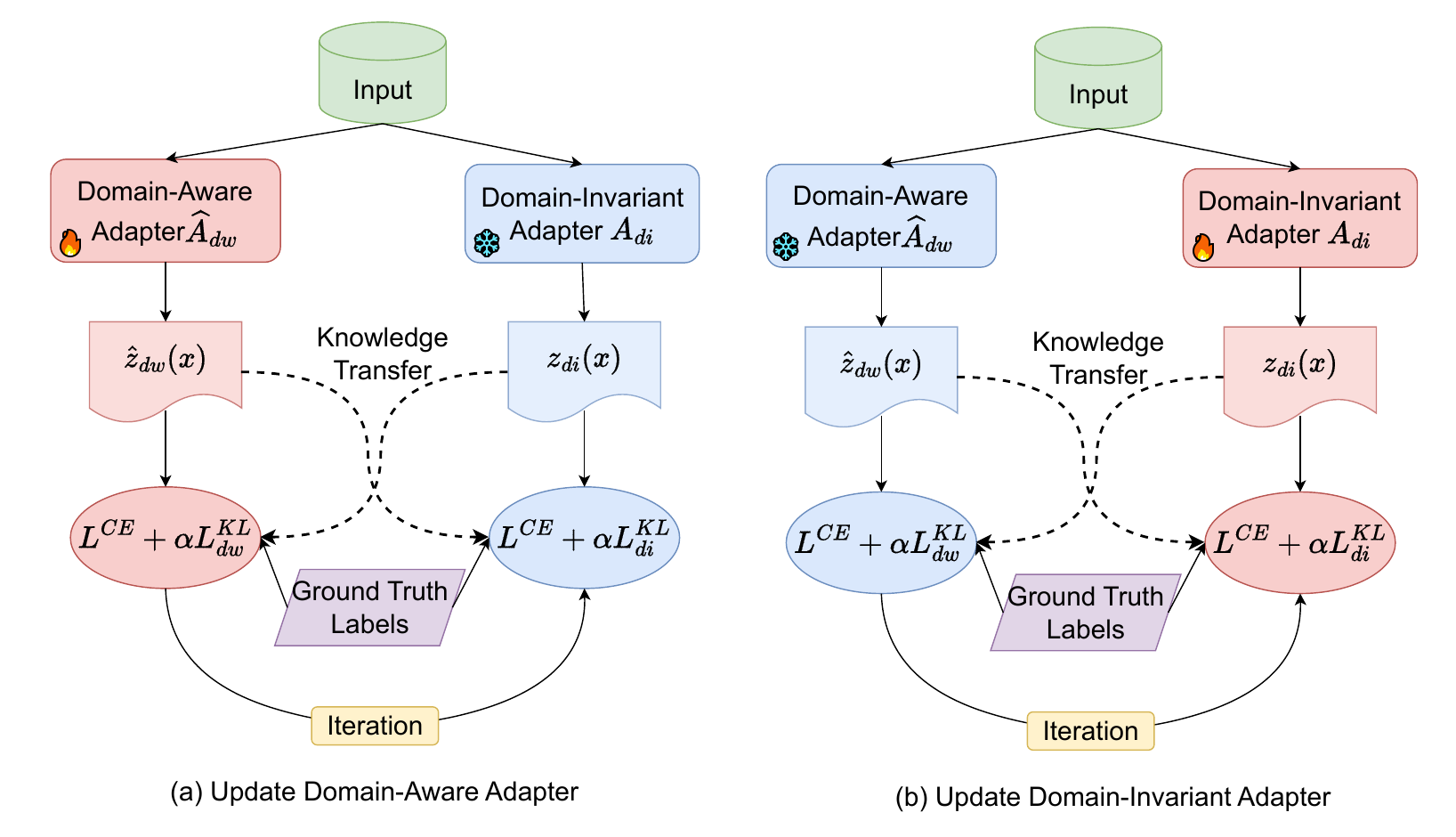}
  \caption{Optimization process.}
  \label{fig:optimization}
\end{figure}

\subsubsection{Bidirectional Knowledge Distillation (BKD)}

The bidirectional knowledge distillation process serves as the core mechanism within the FedSDAF framework, designed to facilitate effective knowledge sharing among various clients. This process enables the model to transfer and integrate knowledge across different source domains, thereby achieving a comprehensive utilization of knowledge from multiple sources. Specifically, the bidirectional knowledge distillation involves computing two Kullback-Leibler (KL) divergences using the following formulas:
\begin{equation}
L_{di}^{KL} = \text{KL}(z_{di}(x) \parallel \hat{z}_{dw}(x))
\end{equation}

\begin{equation}
L_{dw}^{KL} = \text{KL}(\hat{z}_{dw}(x) \parallel z_{di}(x))
\end{equation}
Here, KL denotes the Kullback-Leibler divergence, while $z_{di}$ and $\hat{z}_{dw}$ represent the prediction logits from the model injected with $A_{di}$ and $\hat{A}_{dw}$, respectively. Through this setup, bidirectional knowledge distillation allows clients to exchange knowledge, facilitating learning and adaptation across different source domains. By sharing prediction logits, the model is able to acquire additional domain information from the server pertaining to other domains. This interaction of information is crucial in compensating for knowledge gaps arising from data isolation.

During the actual training process, we employ a distillation loss in conjunction with Cross-Entropy Loss ($L^{CE}$), introducing a weighting factor $\alpha$ to maintain a balance between the two losses. Figure \ref{fig:optimization} illustrates this process of bi-directional distillation:
\begin{equation}
L = L^{CE} + \alpha (L_{di}^{KL} + L_{dw}^{KL})
\end{equation}

\subsubsection{Optimization process}
 The optimization process unfolds across the server and clients. On the server, the global domain-invariant adapter $A_{di}^0$ is initialized. In each communication round $r$, the server distributes the current global adapter $A_{di}^r$ to a subset of clients. Each participating client $k$ receives $A_{di}^r$ and performs local training on its dataset $\mathcal{D}_k$. During local training, the client updates its local copy of the domain-invariant adapter $\hat{A}_{di}^{k,r}$, and its private domain-aware adapter $A_{dw}^k$, using the bidirectional knowledge distillation loss described in Equation (5). After local training, each client sends its updated adapter $\hat{A}_{di}^{k,r}$ back to the server. The server then aggregates the received adapters to produce the next round's global adapter $A_{di}^{r+1}$, using standard federated averaging:

\begin{equation}
    A_{di}^{r+1} \leftarrow \sum_{k=1}^{K} \frac{n_k}{N} \hat{A}_{di}^{k,r}
\end{equation}

where $n_k$ is the number of samples on client $k$ and $N = \sum_k n_k$ is the total number of samples across all clients. This iterative process allows the framework to build a robust global consensus within $A_{di}$ while leveraging the rich, localized knowledge preserved within each client's $A_{dw}$.

The complete training process of FedSDAF unfolds iteratively across the server and clients, as formalized in Algorithm \ref{alg:fedsdaf}.

\begin{algorithm}[htbp]
\caption{Federated Domain Generation with Bidirectional Distillation (FedSDAF)}
\label{alg:fedsdaf}
\begin{algorithmic}[1]
\STATE \textbf{Require:} $K$: Number of clients; $\{D_k\}_{k=1}^K$: Client datasets; $R$: Communication rounds
\STATE \textbf{Ensure:} $\{A_{dw}^{k,R}\}_{k=1}^K$: Well-trained client models; $A_{di}^R$: Global domain-invariant adapter
\STATE \ 
\STATE \textbf{Server Execution:}
\STATE Initialize $A_{di}^0$, $A_{dw}^{k,0} \gets$ random initialization
\STATE Collect $\{n_k = |D_k|\}_{k=1}^K$ from clients
\STATE $N \gets \sum_{k=1}^K n_k$
\STATE $w_k \gets n_k / N,\ \forall k \in \{1,\ldots,K\}$  

\FOR{$r \gets 0$ \textbf{to} $R-1$}
  \IF{$r > 0$}
    \STATE Collect $\{A_{di}^{k,r}\}_{k=1}^K$  
    \STATE $A_{di}^{r+1} \gets \sum_{k=1}^K w_k \cdot A_{di}^{k,r}$
  \ELSE
    \STATE $A_{di}^{r+1} \gets A_{di}^0$
  \ENDIF
  \STATE Broadcast $A_{di}^{r+1}$
\ENDFOR

\STATE \ 
\STATE \textbf{Client $k$ Execution (Parallel):}
\STATE Send $n_k = |D_k|$ to server  
\FOR{$r \gets 0$ \textbf{to} $R-1$}
  \STATE Receive $A_{di}^r$
  \STATE $\hat{A}_{di}^{k,r} \gets A_{di}^r$
  \WHILE{not converged}
    \STATE $\min_{A_{dw}^{k,r}, \hat{A}_{di}^{k,r}} L$ using Equation (5) on $D_k$
  \ENDWHILE
  \STATE Send $A_{di}^{k,r}$
\ENDFOR
\end{algorithmic}
\end{algorithm}

This iterative process allows the framework to build a robust global consensus within the domain-invariant adapter $A_{di}$ while continuously enriching it with the diverse, localized knowledge preserved within each client's private domain-aware adapter $A_{dw}$.

\section{Experimental Results}
In this section, we describe the details of our experimental setup and report on a series of experimental results to illustrate the validity of our approach.

\subsection{Datasets and Baselines}
\noindent\textbf{Datasets.} We evaluate our method on four standard DG benchmarks: \textbf{PACS} \cite{8237853}, \textbf{OfficeHome} \cite{8100055}, \textbf{VLCS} \cite{6751316}, and the large-scale \textbf{DomainNet} \cite{9010750}. For all datasets, we employ a leave-one-domain-out evaluation protocol, where one domain is held out as the unseen target for testing.

\noindent\textbf{Baseline Methods.} We compare FedSDAF against 14 methods from three categories:
\begin{enumerate}
    \item \textbf{Centralized DG Methods:} These methods require centralized data access. Baselines include \textbf{SWAD} \cite{10.5555/3540261.3541977}, \textbf{HCVP} \cite{10814670}, and \textbf{Doprompt} \cite{zheng2022promptvisiontransformerdomain}.
    \item \textbf{CNN-based FedDG Methods:} These methods use CNN backbones in a federated setting. Baselines include \textbf{FedSR} \cite{10.5555/3600270.3603084}, \textbf{FedADG} \cite{zhang2023federatedlearningdomaingeneralization}, \textbf{CCST} \cite{10030422}, \textbf{ELCFS} \cite{9577482}, and \textbf{GA} \cite{10203192}.
    \item \textbf{Parameter-Efficient Fine-Tuning (PEFT) Methods:} These recent methods are most comparable to our approach. Baselines include \textbf{FedCLIP} \cite{Lu2023FedCLIPFG}, \textbf{PromptFL} \cite{10210127}, \textbf{FedAPT} \cite{10.1609/aaai.v38i13.29434}, \textbf{FedPR} \cite{10205077}, \textbf{FedMaPLe} \cite{10203359}, and \textbf{PLAN} \cite{gong2024federateddomaingeneralizationprompt}.
\end{enumerate}

\begin{table*}[t]
\centering
\caption{Performance Comparison on PACS, VLCS, and OfficeHome Datasets (\%)}
\label{tab:combined}
\setlength{\tabcolsep}{4pt} 
\small 
\begin{tabular}{@{}l S[table-format=2.2]S[table-format=2.2]S[table-format=2.2]S[table-format=2.2] S[table-format=2.2] S[table-format=2.2]S[table-format=2.2]S[table-format=2.2]S[table-format=2.2] S[table-format=2.2] S[table-format=2.2]S[table-format=2.2]S[table-format=2.2]S[table-format=2.2] S[table-format=2.2]@{}}
\toprule
\multirow{2}{*}{\textbf{Methods}} & \multicolumn{5}{c}{\textbf{PACS}} & \multicolumn{5}{c}{\textbf{VLCS}} & \multicolumn{5}{c}{\textbf{OfficeHome}} \\
\cmidrule(lr){2-6} \cmidrule(lr){7-11} \cmidrule(lr){12-16}
& {A} & {C} & {P} & {S} & {\textbf{Avg.}} & {C} & {L} & {V} & {S} & {\textbf{Avg.}} & {A} & {C} & {P} & {R} & {\textbf{Avg.}} \\
\midrule

\multicolumn{16}{@{}l}{\textit{Centralized learning based Domain Generalization Methods}} \\
SWAD       & 93.23 & 85.93 & 99.18 & 82.03 & 90.44 & 98.49 & 68.36 & 75.40 & 79.49 & 79.31 & 76.26 & 68.87 & 86.74 & 87.03 & 79.73 \\
HCVP       & 93.17 & 86.89 & 99.33 & 81.30 & 90.17 & 96.32 & 66.26 & 76.40 & 81.65 & 81.08 & 81.77 & 69.76 & 88.01 & 90.62 & 82.54 \\
Doprompt   & 95.00 & 86.35 & 99.63 & 78.20 & 89.91 & 96.70 & 66.53 & 78.28 & 79.39 & 80.23 & 80.95 & 70.88 & 88.94 & 90.10 & 82.72 \\
\addlinespace[0.5em]

\multicolumn{16}{@{}l}{\textit{CNN-based FedDG Methods (backbone: Resnet-50)}} \\
FedSR      & 88.19 & 67.45 & 95.74 & 65.92 & 79.33 & 95.16 & 65.86 & 78.51 & 73.49 & 78.26 & 69.12 & 49.69 & 72.71 & 79.12 & 67.66 \\
FedADG     & 82.93 & 65.42 & 98.09 & 65.36 & 77.95 & 95.21 & 65.76 & 76.43 & 75.96 & 78.34 & 69.31 & 48.76 & 72.89 & 79.13 & 67.52 \\
CCST       & 87.02 & 74.57 & 98.29 & 65.84 & 81.43 & 96.49 & 65.73 & 76.42 & 77.67 & 79.08 & 69.23 & 51.36 & 72.09 & 81.27 & 68.19 \\
ELCFS      & 86.77 & 73.21 & 98.14 & 65.16 & 80.82 & 95.67 & 65.02 & 76.55 & 77.96 & 79.80 & 68.17 & 50.52 & 71.44 & 80.11 & 67.56 \\
GA         & 87.68 & 75.19 & 97.56 & 65.86 & 81.57 & 96.77 & 65.16 & 78.89 & 78.93 & 79.18 & 68.62 & 50.60 & 73.35 & 81.23 & 68.45 \\
\addlinespace[0.5em]

\multicolumn{16}{@{}l}{\textit{PEFT Methods (backbone: ViT-B/16)}} \\
FedCLIP    & 96.19 & 97.91 & 99.76 & 85.85 & 94.93 & 99.93 & 66.98 & 73.28 & 87.14 & 81.83 & 78.45 & 64.77 & 73.28 & 87.84 & 79.69 \\
PromptFL   & 96.34 & 98.46 & 99.58 & 92.19 & 96.64 & 99.71 & 68.03 & 72.24 & 85.10 & 83.59 & 82.98 & 68.98 & 92.14 & 90.27 & 83.59 \\
FedAPT     & 97.15 & 99.12 & 99.69 & 92.34 & 97.08 & 99.36 & 68.18 & 81.06 & 85.98 & 83.64 & 83.96 & 71.65 & 91.93 & 90.51 & 84.51 \\
FedPR      & 98.10 & 99.02 & 99.88 & 91.11 & 97.03 & 99.36 & 68.18 & 81.06 & 85.98 & 83.64 & 84.04 & 71.63 & 92.39 & 91.34 & 84.58 \\
FedMaPLe   & 98.44 & 99.02 & 99.94 & 90.40 & 96.95 & 98.02 & 69.54 & 82.15 & 85.81 & 83.87 & 84.56 & 72.82 & 92.38 & 91.07 & 85.21 \\
PLAN       & 98.58 & 99.14 & \textbf{99.82} & 92.08 & 97.40 & \textbf{99.18} & 69.94 & 83.75 & 88.28 & 85.29 & 86.65 & 74.73 & 93.47 & 92.06 & 86.73 \\
\midrule
\textbf{FedSDAF} & \textbf{98.88} & \textbf{99.36} & 99.28 & \textbf{96.56} & \textbf{98.52} & 98.72 & \textbf{85.09} & \textbf{87.38} & \textbf{94.33} & \textbf{91.38} & \textbf{87.15} & \textbf{76.20} & \textbf{94.38} & \textbf{92.78} & \textbf{87.63} \\
\bottomrule
\end{tabular}
\end{table*}

\begin{table}[htbp]
\centering
\caption{Performance Comparison on DomainNet Datasets (\%)}
\label{tab:domainnet}
\begingroup
\small 
\setlength{\tabcolsep}{3pt} 
\begin{tabular}{@{}l*{7}{S[table-format=2.2]}@{}}
\toprule
\multirow{2}{*}{\textbf{Methods}} & \multicolumn{7}{c}{\textbf{DomainNet}} \\
\cmidrule(lr){2-8}
 & {C} & {I} & {P} & {Q} & {R} & {S} & {\textbf{Avg.}} \\
\midrule
FedCLIP    & 74.12 & 48.36 & 68.49 & 31.73 & 80.52 & 58.62 & 60.31 \\
PromptFL   & 76.53 & 51.72 & 70.86 & 34.21 & 81.68 & 68.37 & 63.90 \\
FedAPT     & 77.02 & 51.45 & 70.36 & 49.62 & 86.64 & 68.43 & 67.25 \\
FedPR      & 75.49 & 51.96 & 71.42 & 35.98 & 82.67 & 69.43 & 64.49 \\
FedMaPLe   & 78.61 & 65.23 & 71.89 & 43.46 & 86.32 & 72.46 & 69.67 \\
PLAN       & 79.51 & \textbf{66.42} & 72.11 & 48.83 & \textbf{86.72} & 72.69 & 71.05 \\
\midrule
\textbf{FedSDAF} & \textbf{82.59} & 62.95 & \textbf{85.12} & \textbf{54.16} & 84.99 & \textbf{74.14} & \textbf{73.99} \\
\bottomrule
\end{tabular}
\endgroup
\end{table}

\subsection{Implementation Details}

In our implementation of FedSDAF, we employed the ViT-B16 architecture as the backbone network. The federated learning process was configured to span 200 communication rounds, with each client performing 30 local epochs per round. To accommodate the diverse characteristics of the benchmark datasets, we utilized dataset-specific batch sizes: 128 for both the PACS and OfficeHome datasets, 64 for VLCS, and 1024 for the more extensive DomainNet. For optimization, we adopted the Scaffold optimizer with an initial learning rate of 0.001 to facilitate effective coordination of client model updates. To regulate the learning rate throughout the training process, we implemented PyTorch's StepLR scheduler with a step size of 0.2 and a decay factor of 0.1, ensuring appropriate learning rate adjustments during model convergence.

\subsection{Main Results}
Tables \ref{tab:combined} and \ref{tab:domainnet} show the comparison of our proposed FedSDAF method with other methods on PACS, VLCS, OfficeHome, and DomainNet, respectively.

In our experimental evaluations, we achieved state-of-the-art (SOTA) results across all four mainstream datasets: PACS, VLCS, OfficeHome, and DomainNet. This accomplishment underscores the robust generalization capabilities of our proposed Federated Source Domain Awareness Framework (FedSDAF). Notably, our method demonstrated significant advancements on the Sketch domain of the PACS dataset, where previous approaches had struggled to deliver satisfactory outcomes. Specifically, our framework achieved an impressive accuracy of 96.56\%, reflecting a notable improvement of 4.22\% over prior methods.

Furthermore, we observed strong performance on DomainNet, particularly in the Quickdraw domain, where our method excelled despite the inherent challenges posed by unnatural image features. This success further validates our framework's adaptability to diverse and complex data distributions. The VLCS dataset, which integrates images from varying photographic styles, highlights the differences in visual representation that our model adeptly addressed. By effectively leveraging domain-aware features from the source domains, our method capitalized on the nuances of each dataset, thereby enhancing its performance across the board.

Standard Domain Generalization (DG) methods, while demonstrating excellent performance across various datasets, rely on centralized frameworks that necessitate unrestricted access to client data, which compromises data privacy. This centralized approach facilitates robust data management systems but raises significant concerns regarding data protection. In contrast, our FedSDAF framework addresses these privacy concerns by leveraging a federated learning paradigm, which allows for the exchange of knowledge without compromising client data.

Existing Federated Domain Generalization (FedDG) methods, such as FedSR and FedADG, exhibit limitations, particularly in their reliance on local data without sufficient integration of domain-specific knowledge from diverse sources. While these methods have made strides in representation learning, they often fall short in achieving robust generalization across unseen domains. In contrast, our framework incorporates the utilization of source-domain perceptual features, which have proven instrumental in enhancing model adaptability to unseen target domains. This is particularly evident in our results, where the integration of domain-aware features has led to significant performance gains. Additionally, our bidirectional knowledge distillation mechanism facilitates effective knowledge sharing between clients, thereby enhancing model adaptability.

\subsection{Ablation Experiment}

\begin{table}[htbp]
  \centering
  \caption{Ablation Study for Different Components}
  \label{tab:components}
  \begingroup
  \small
  \setlength{\tabcolsep}{5pt}
  \begin{tabular}{ccccccc}
    \toprule
    BKD & MHSA & $A_{di}$ & $A_{dw}$ & PACS & OfficeHome & VLCS \\
    \midrule
    \checkmark & \checkmark & \checkmark & \checkmark & 98.52 & 87.63 & 91.38 \\
    $\times$     & \checkmark & \checkmark & \checkmark & 97.46 & 85.81 & 86.19 \\
    \checkmark & $\times$     & \checkmark & \checkmark & 97.13 & 85.42 & 87.20 \\
    \checkmark & $\times$     & \checkmark & $\times$     & 95.47 & 83.60 & 83.98 \\
    $\times$     & $\times$     & \checkmark & $\times$     & 95.46 & 82.17 & 83.51 \\
    $\times$     & $\times$     & $\times$     & $\times$     & 93.29 & 78.70 & 82.69 \\
    \bottomrule
  \end{tabular}
  \endgroup
\end{table}

To illustrate the significance of the various components utilized in the Federated Source Domain Awareness Framework (FedSDAF), we conducted an ablation study across three benchmark datasets. The results are presented in Table \ref{tab:components}.

Optimal model performance was attained when all components were active, demonstrating the synergistic integration of features that collectively enhance the model's generalization capacity. In contrast, disabling both the BKD and the MHSA led to a significant decline in performance, highlighting the critical roles these components play in facilitating information exchange and feature extraction. The observed drop in accuracy indicates that, without these mechanisms, the model struggles to maintain the understanding necessary for effective generalization. This finding underscores the necessity of collaborative learning strategies, such as BKD, in optimizing model performance within federated contexts characterized by data isolation.

Further analysis revealed that disabling either adapter resulted in a decrease in model accuracy, emphasizing their integral roles within the architecture. Specifically, the reduction in performance upon removing the domain-invariant adapter underscores its importance in extracting stable features crucial for robustness against domain shifts. The lowest performance was recorded when all components were disabled, confirming the complementary and synergistic nature of our framework. This substantial reduction in accuracy further emphasizes the necessity of each component's functionality within the overall system.

\subsection{Convergence Analysis}

\begin{figure}[htbp]
  \centering
  \includegraphics[width=\columnwidth]{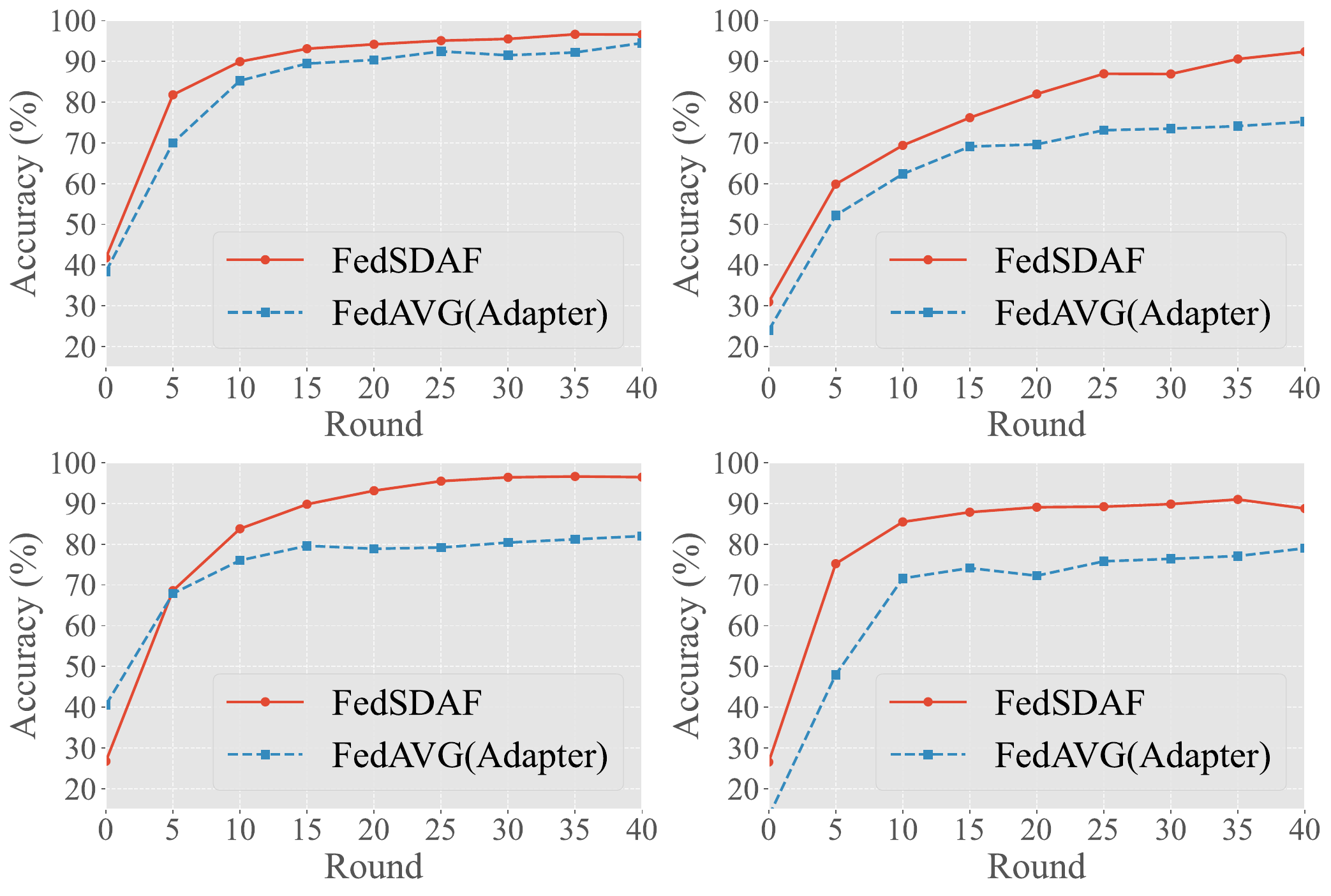}
  \caption{Convergence analysis on different clients in domain benchmark.}
  \label{fig:convergence}
\end{figure}

To validate the convergence of our FedSDAF framework, we compared it against the FedAvg (Adapter) baseline on the PACS dataset. As shown in Figure \ref{fig:convergence}, FedSDAF converges significantly faster and more stably. Within the first 10 rounds, FedSDAF already achieves a 5.8--7.2\% accuracy advantage. This rapid start is driven by the local domain-aware adapter's MHSA mechanism, which accelerates the extraction of domain-specific features, whereas FedAvg's domain-agnostic aggregation lags behind. By round 40, FedSDAF's lead widens to a stable 9.4--13.8\% accuracy margin. This superior stability is a direct result of our bidirectional knowledge distillation, which systematically fuses complementary knowledge across domains. This dynamic process overcomes the limitations of static aggregation strategies that are typically hindered by domain heterogeneity, ensuring a consistently faster and more robust convergence.

\subsection{Hyperparameter Analysis}

\begin{figure}[htbp]
  \centering
  \includegraphics[width=\columnwidth]{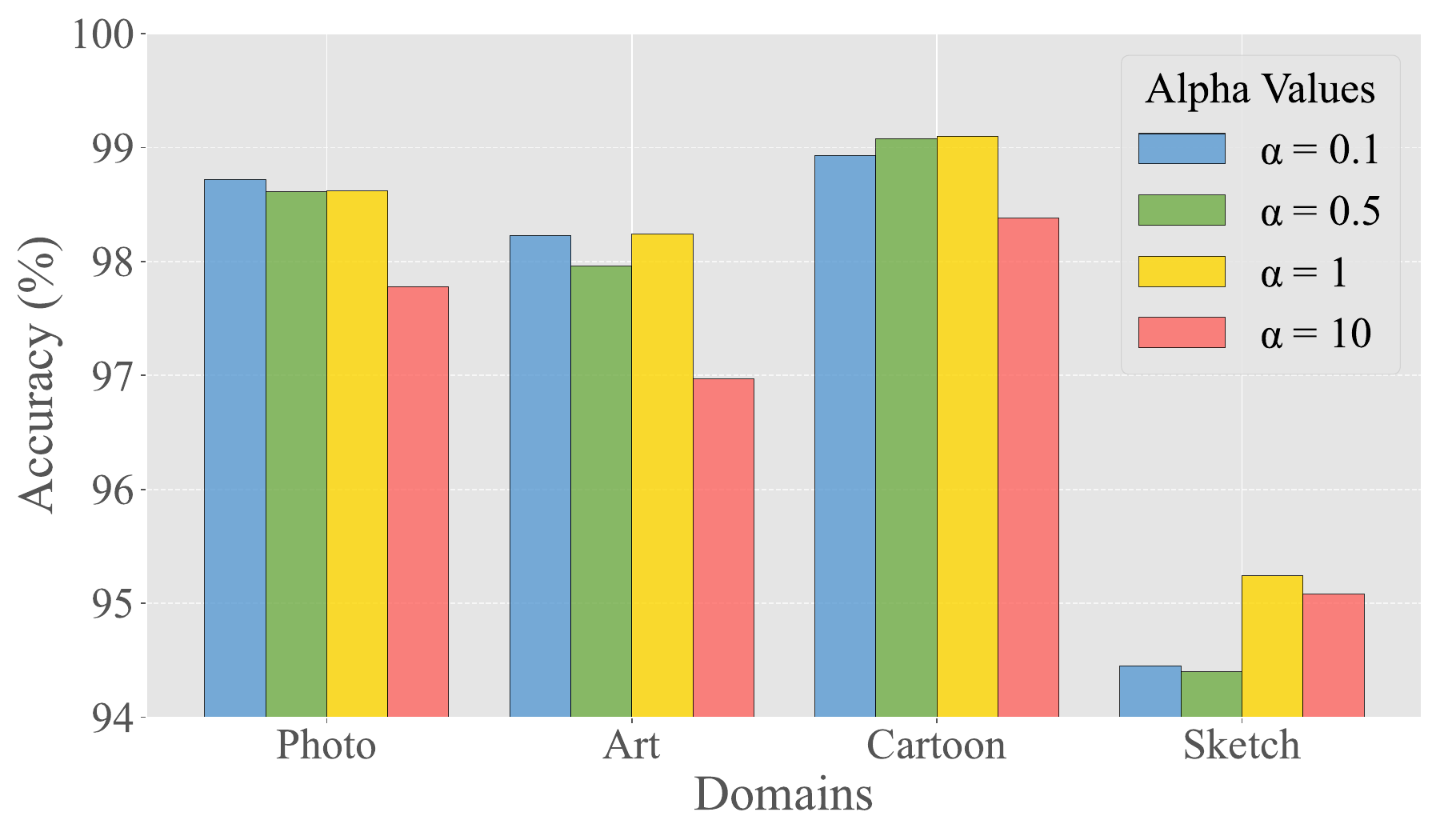}
  \caption{Effects of $\alpha$.}
  \label{fig:alpha_sensitivity}
\end{figure}

We evaluate the robustness of FedSDAF through systematic analysis of the loss balancing coefficient $\alpha$ (Eq. 5) across the entire PACS benchmark. As shown in Figure \ref{fig:alpha_sensitivity}, FedSDAF consistently achieves accuracy exceeding 97\% when $\alpha$ ranges from 0.3 to 5, with minimal variance ($\pm$0.8\%). Notably, even with $\alpha=0.1$ (representing reduced distillation influence), the model maintains 96.2\% accuracy, demonstrating the inherent stability of our dual-adapter architecture. Optimal performance is observed at $\alpha=1$, where domain-specific knowledge preservation and cross-client fusion reach an equilibrium. When $\alpha$ exceeds 5, a slight performance decline occurs (97.1\% at $\alpha=10$), as the over-strengthened global distillation signal marginally inhibits local adaptation capabilities. These experimental results confirm FedSDAF's robustness to hyperparameter selection, requiring only the straightforward choice of $\alpha=1$ to achieve peak performance without extensive tuning.

\subsection{Communication Cost Analysis}
A concern regarding our framework could be the complexity introduced by the MHSA mechanism. However, this design choice does not translate to increased communication overhead. The Domain-Aware Adapter, which contains the MHSA module, is kept strictly local to each client and is never aggregated by the server. Therefore, its parameters do not contribute to the communication cost.

\begin{figure}[htbp]
     \centering
     \includegraphics[width=\columnwidth]{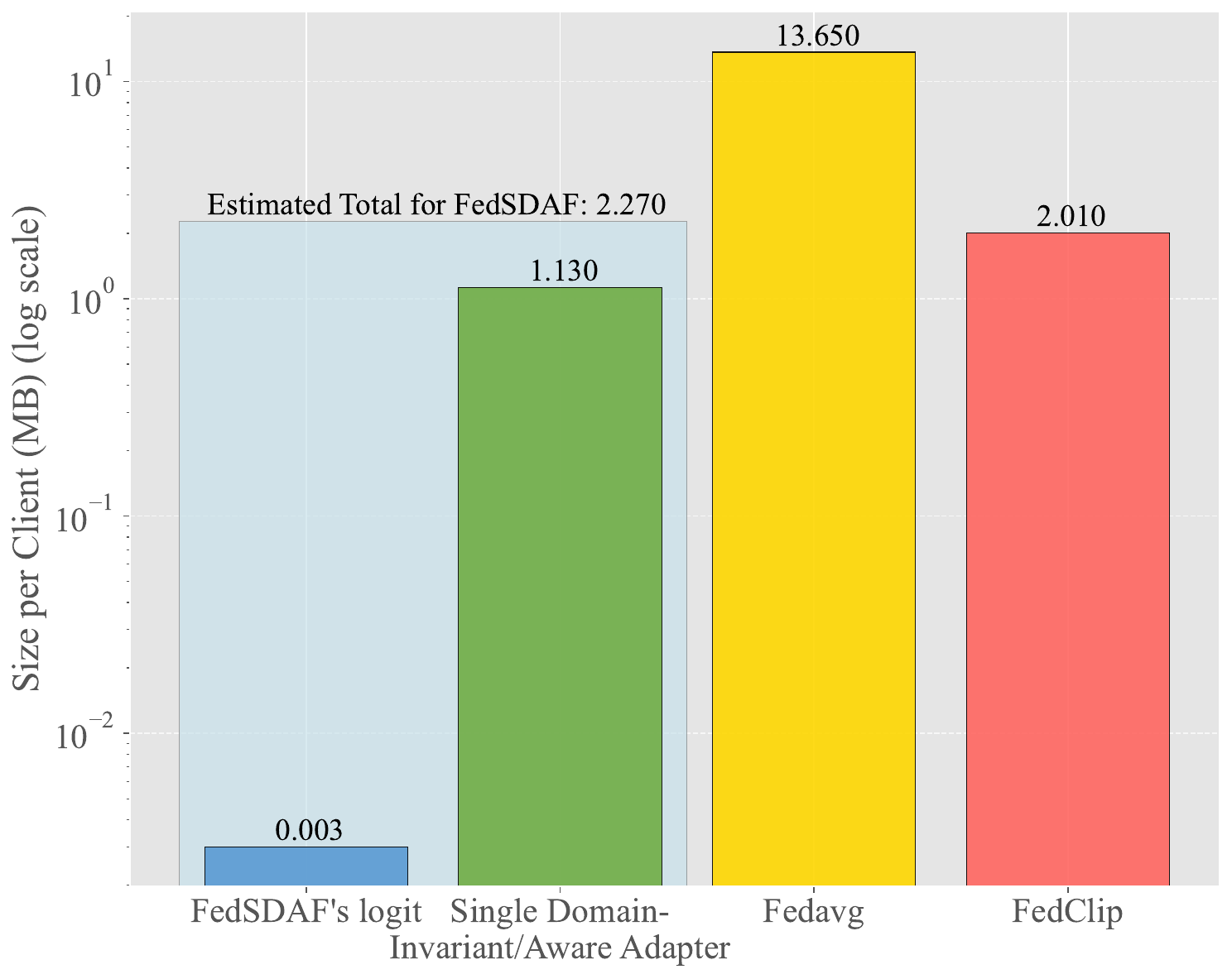}
     \caption{Communication cost comparison per client. FedSDAF's total cost is comparable to other adapter-based methods and significantly lower than full-model federation.}
     \label{fig:comm_cost}
\end{figure}

The communication cost per round is dominated by two components: the upload/download of the lightweight domain-invariant adapter $\hat{A}_{di}$, and the exchange of logits for the BKD process. As shown in Figure~\ref{fig:comm_cost}, the size of the logits is negligible (0.003 MB). The total estimated communication cost for FedSDAF per client per round is approximately 2.270 MB. This is not only comparable to other PEFT-based methods like FedCLIP (2.010 MB) but is also an order of magnitude smaller than full-model aggregation methods like FedAvg (13.650 MB). This analysis demonstrates that FedSDAF remains highly communication-efficient, highlighting its practical potential for deployment in real-world federated learning scenarios.

\subsection{Further Analysis}

\subsubsection{Fine-grained Ablation Study on Hard Domains}
A potential concern is that the performance gain from components like MHSA appears modest when averaged across all domains. This is because performance on many domains is already saturated (e.g., >99\% accuracy), masking the components' true impact. To provide a clearer picture, we conduct a fine-grained ablation study focusing on the most challenging domains from three benchmarks: PACS (Photo), VLCS (LabelMe), and OfficeHome (Clipart), where models typically struggle the most.

The results are presented in Table~\ref{tab:hard_domain_ablation}. On these hard domains, the contribution of each component becomes significantly more pronounced. For instance, enabling BKD and MHSA leads to substantial gains of 3--7\% on VLCS and OfficeHome. The full model consistently achieves the best results, with the MHSA mechanism being particularly crucial for boosting performance on these challenging generalization tasks. This underscores the importance of each component in our framework for achieving robust generalization, an effect that is most visible where it matters most.

\begin{table}[htbp]
\centering
\caption{Fine-grained ablation study on the most challenging domains. The performance improvements for each component are more significant on these domains compared to the overall average.}
\label{tab:hard_domain_ablation}
\footnotesize
\setlength{\tabcolsep}{3pt}
\begin{tabular}{ccccccc}
\toprule
\multicolumn{4}{c}{\textbf{Components}} & \multicolumn{3}{c}{\textbf{Accuracy (\%) on Hard Domains}} \\
\cmidrule(r){1-4} \cmidrule(l){5-7}
\textbf{BKD} & \textbf{MHSA} & $\boldsymbol{A_{di}}$ & $\boldsymbol{A_{dw}}$ & \textbf{PACS (P)} & \textbf{VLCS (L)} & \textbf{OfficeHome (C)} \\
\midrule
\checkmark & \checkmark & \checkmark & \checkmark & \textbf{99.28} & \textbf{85.09} & \textbf{76.20} \\
\midrule
$\times$ & \checkmark & \checkmark & \checkmark & 98.53 & 82.15 & 73.45 \\
\checkmark & $\times$ & \checkmark & \checkmark & 98.21 & 81.88 & 72.93 \\
\checkmark & \checkmark & $\times$ & \checkmark & 97.55 & 79.50 & 70.10 \\
\checkmark & \checkmark & \checkmark & $\times$ & 97.98 & 80.20 & 71.30 \\
\midrule
$\times$ & $\times$ & $\times$ & $\times$ & 95.10 & 74.30 & 68.50 \\
\bottomrule
\end{tabular}
\end{table}

\subsubsection{Visualization}

\begin{figure}[htbp]
  \centering
  \includegraphics[width=\columnwidth]{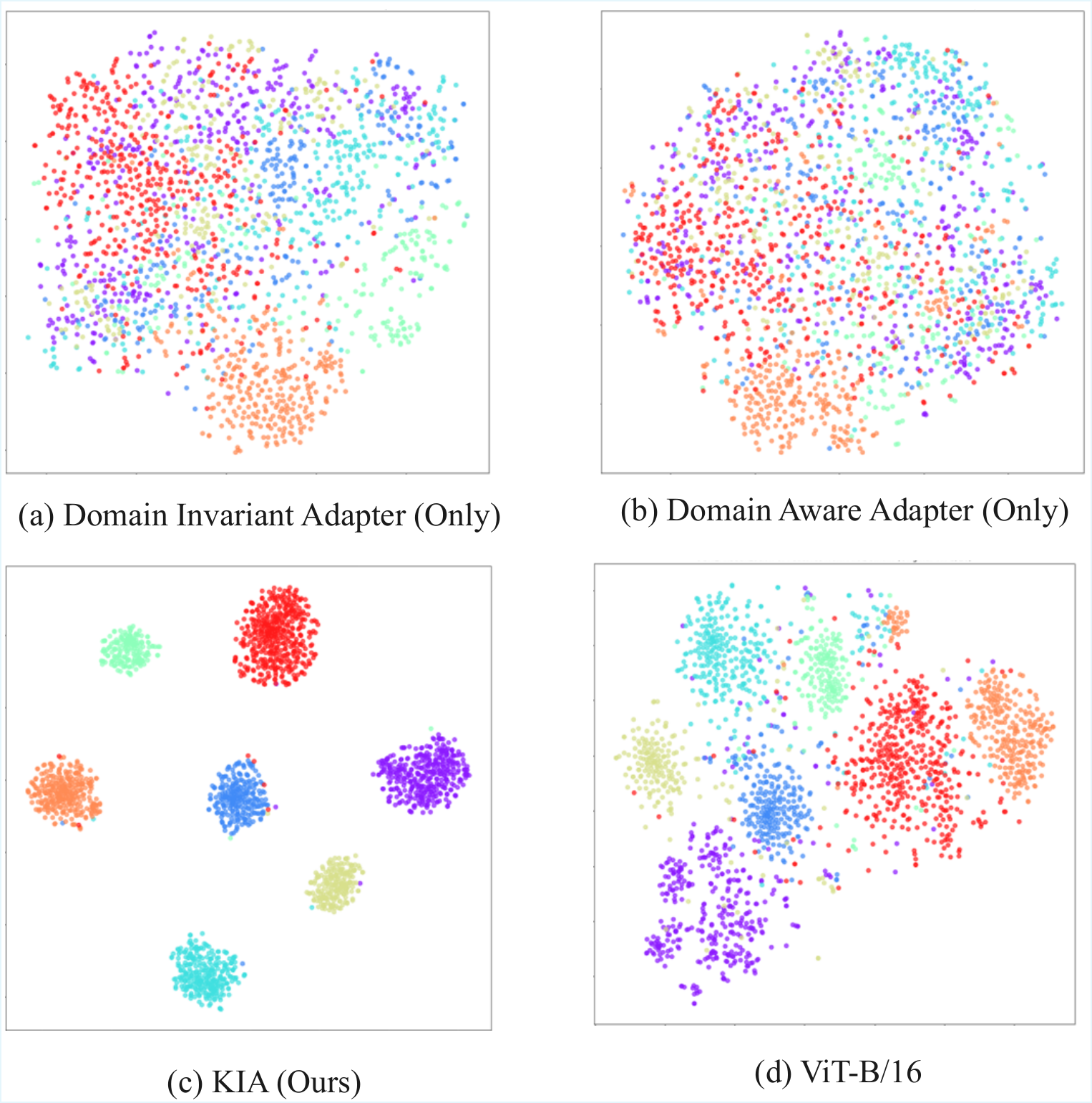}
  \caption{Visualization of the target features extracted by ViT and FedSDAF.}
  \label{fig:tsne}
\end{figure}

We present several visualization results to provide an intuitive understanding of the advantages of FedSDAF. In Fig. \ref{fig:tsne}, we apply the t-SNE algorithm to visualize the features of target samples extracted by ViT-B/16, the Domain-Invariant Adapter, the Domain-Aware Adapter, and FedSDAF (KIA), respectively, with Sketch designated as the target domain on PACS. The t-SNE visualization reveals critical insights into FedSDAF's feature discrimination capability within the challenging Sketch domain. While ViT-B/16 (Fig. \ref{fig:tsne}(d)) exhibits a dispersed feature distribution with substantial inter-class overlap, both individual adapters demonstrate distinct adaptation strategies aligned with our design motivation. The domain-invariant adapter $\hat{A}_{di}$ (Fig. \ref{fig:tsne}(a)) produces elliptical feature clusters with overlapping boundaries, reflecting its emphasis on cross-domain alignment at the expense of category discrimination. Conversely, the domain-aware adapter $\hat{A}_{dw}$ (Fig. \ref{fig:tsne}(b)) generates compact but partially misaligned clusters, indicative of its focus on capturing source-specific discriminative features while neglecting domain shift mitigation.
FedSDAF (Fig. \ref{fig:tsne}(c)) fundamentally overcomes these limitations through hierarchical feature disentanglement. The isolated red cluster in the lower-left quadrant manifests the domain-aware component's success in preserving source-discriminative patterns, while the uniform spacing between blue and purple clusters in the upper region verifies the domain-invariant module's effectiveness in establishing cross-domain consistency. This dual mechanism achieves geometric synergy: the domain-aware adapter $\hat{A}_{dw}$ actively pushes class clusters apart by amplifying source-specific discriminators, while the domain-invariant adapter $\hat{A}_{di}$ passively pulls features toward domain-agnostic decision boundaries. Such collaborative dynamics explain why FedSDAF surpasses both the base model and standalone adapters, as visualized by the sharpened cluster margins and reduced overlap density compared to other configurations.
These visual patterns collectively validate our core motivation: source-aware feature refinement requires simultaneous discriminative pattern preservation (through domain-aware adaptation) and cross-domain topology alignment (via invariant feature projection), which FedSDAF achieves through its dual-branch architecture.

\subsubsection{Scalability with Increasing Client Pool Size}
\label{sec:scalability}

A critical aspect of any federated learning framework is its ability to scale effectively as the number of participating clients grows. To assess the scalability of FedSDAF, we conducted an experiment on the PACS dataset by varying the number of clients ($K$) among which the source domains are partitioned. We evaluated the framework's performance with client pool sizes of $K=4, 20, 30,$ and $50$.

\begin{figure}[htbp]
    \centering
    \includegraphics[width=\columnwidth]{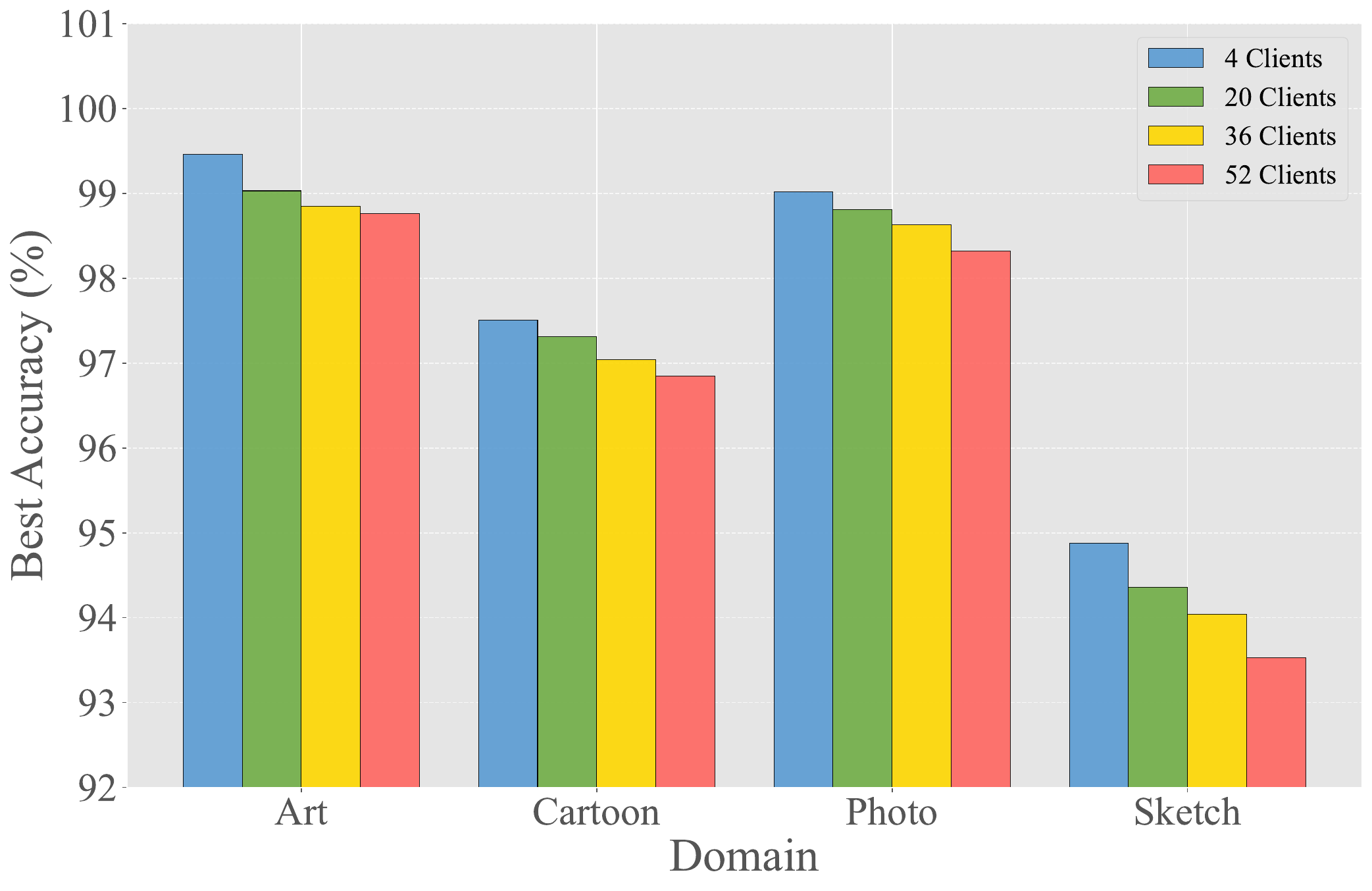}
    \caption{Performance of FedSDAF on the PACS dataset with a varying number of clients ($K$). The results show a graceful and minor decrease in accuracy as the client pool size increases, demonstrating the framework's strong scalability.}
    \label{fig:scalability}
\end{figure}

The results, presented in Figure~\ref{fig:scalability}, demonstrate that FedSDAF exhibits robust scalability. As the number of clients increases, there is a slight and graceful degradation in accuracy across all domains. This trend is expected in federated settings, as a larger client pool typically introduces greater data heterogeneity and makes achieving a global consensus more challenging. However, the performance decline is modest, even when scaling to 50 clients. For instance, on the most challenging Sketch domain, the accuracy remains high, showcasing the framework's ability to maintain its effectiveness. This highlights that FedSDAF's core mechanisms, particularly the synergistic learning between the domain-aware and domain-invariant adapters, are resilient to increased decentralization, making the framework highly suitable for large-scale, real-world deployments.

\subsubsection{Analysis of Adapter Fusion Strategy}
\label{sec:fusion_strategy}

Our framework integrates the outputs of the domain-aware ($A_{dw}$) and domain-invariant ($A_{di}$) adapters using a simple summation with equal weights, as described in Equation (2): $h^{\prime}=h+\frac{1}{2}\hat{A}_{dw}(h)+\frac{1}{2}\hat{A}_{di}(h)$. To validate this design choice, we performed an ablation study to analyze the impact of different fusion strategies on the PACS dataset. We explored two alternatives: modifying the relative weighting of the adapter outputs, and replacing the summation with a concatenation operation.

\begin{figure}[htbp]
    \centering
    \includegraphics[width=\columnwidth]{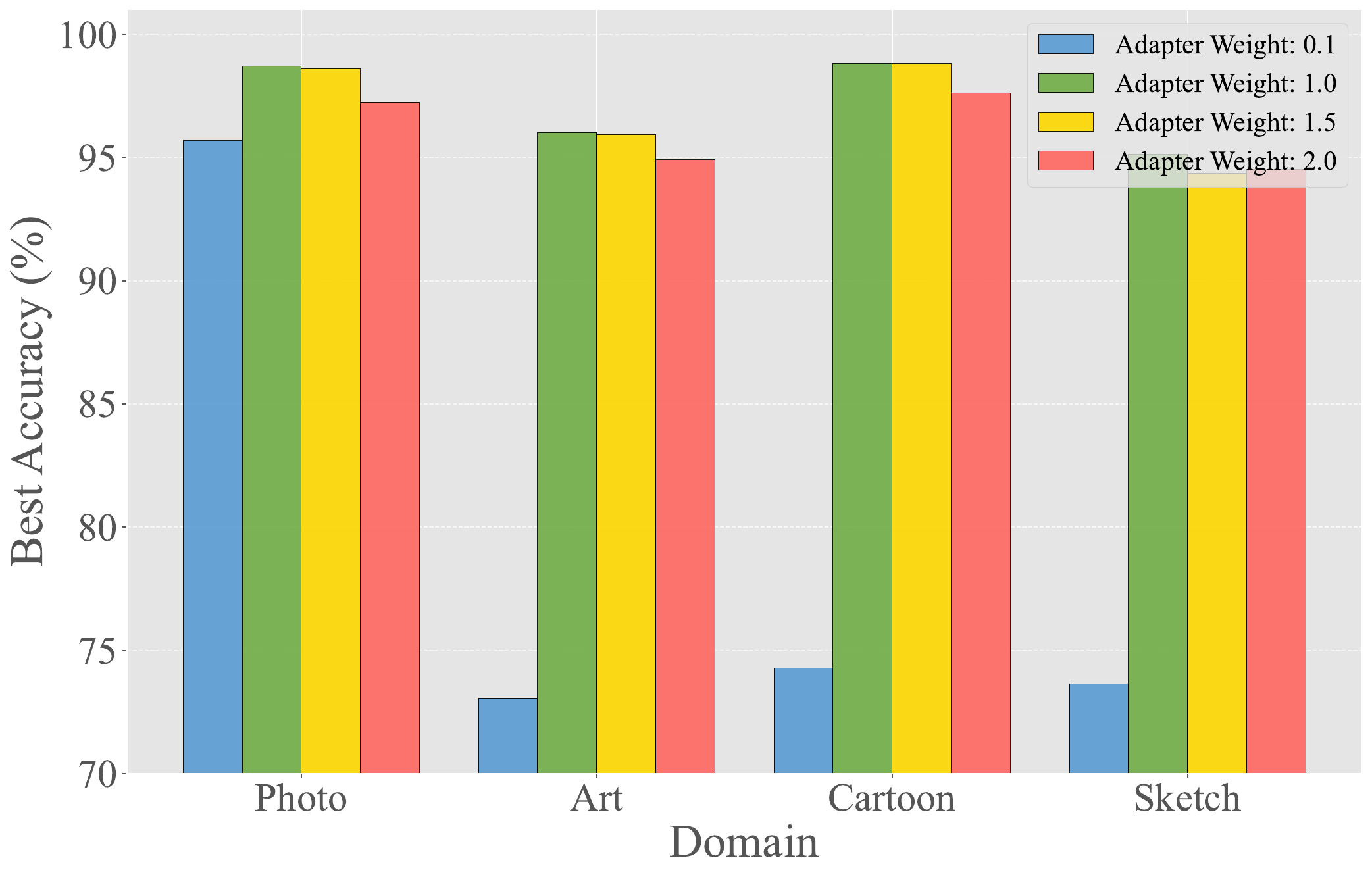}
    \caption{Performance comparison of different adapter fusion strategies on the PACS dataset. Both the weighted-sum and concatenation methods prove effective, with a balanced weighting scheme performing optimally.}
    \label{fig:fusion_strategy}
\end{figure}

The results are illustrated in Figure~\ref{fig:fusion_strategy}. The 'Weight' parameter in the legend corresponds to the balance between the two adapters, where 'Weight=1' represents our original equal-weighting scheme. The analysis reveals several key insights:
\begin{itemize}
    \item \textbf{Robustness to Weighting:} The model achieves the highest performance with 'Weight=1' (our default) and 'Weight=1.5', indicating that a balanced contribution is optimal. Performance degrades significantly when the weight is too low (e.g., 'Weight=0.1'), confirming that a substantial signal from both adapters is crucial for success.
    \item \textbf{Effectiveness of Concatenation:} The 'Concat' strategy, where the feature vectors from both adapters are concatenated, also yields excellent results. Its performance is highly competitive with the optimal weighted-sum approach across all domains.
\end{itemize}

This study confirms that our default equal-weighting scheme is a simple, robust, and effective choice. Furthermore, the strong performance of the concatenation alternative validates the core hypothesis of FedSDAF: the explicit disentanglement and subsequent reintegration of local expertise and global consensus are the primary drivers of performance, and the framework is not overly sensitive to the specific fusion mechanism employed.

\subsubsection{Ablation Study on MHSA via Dynamic Distillation Weighting}
\label{sec:ablation_mhsa}

In our main experiments, the FedSDAF framework utilizes a MHSA mechanism alongside a fixed weight for the BKD loss. To validate the unique contribution of the MHSA component, we conducted an ablation study where we removed it and attempted to compensate by introducing a more sophisticated, dynamic weighting strategy for the distillation loss. Specifically, instead of a fixed $\alpha$, we employed an exponentially increasing weight governed by the formula:
\begin{equation}
    \alpha_{t} = \alpha_{\text{max}} \times \left(1 - e^{-\frac{t}{\tau}}\right)
\end{equation}
where $t$ is the current epoch, $\alpha_{\text{max}}$ is the maximum weight, and $\tau$ controls the rate of increase. The goal was to see if an optimally tuned, dynamic distillation process could replicate the performance gains from MHSA.

\begin{figure}[htbp]
    \centering
    \includegraphics[width=\columnwidth]{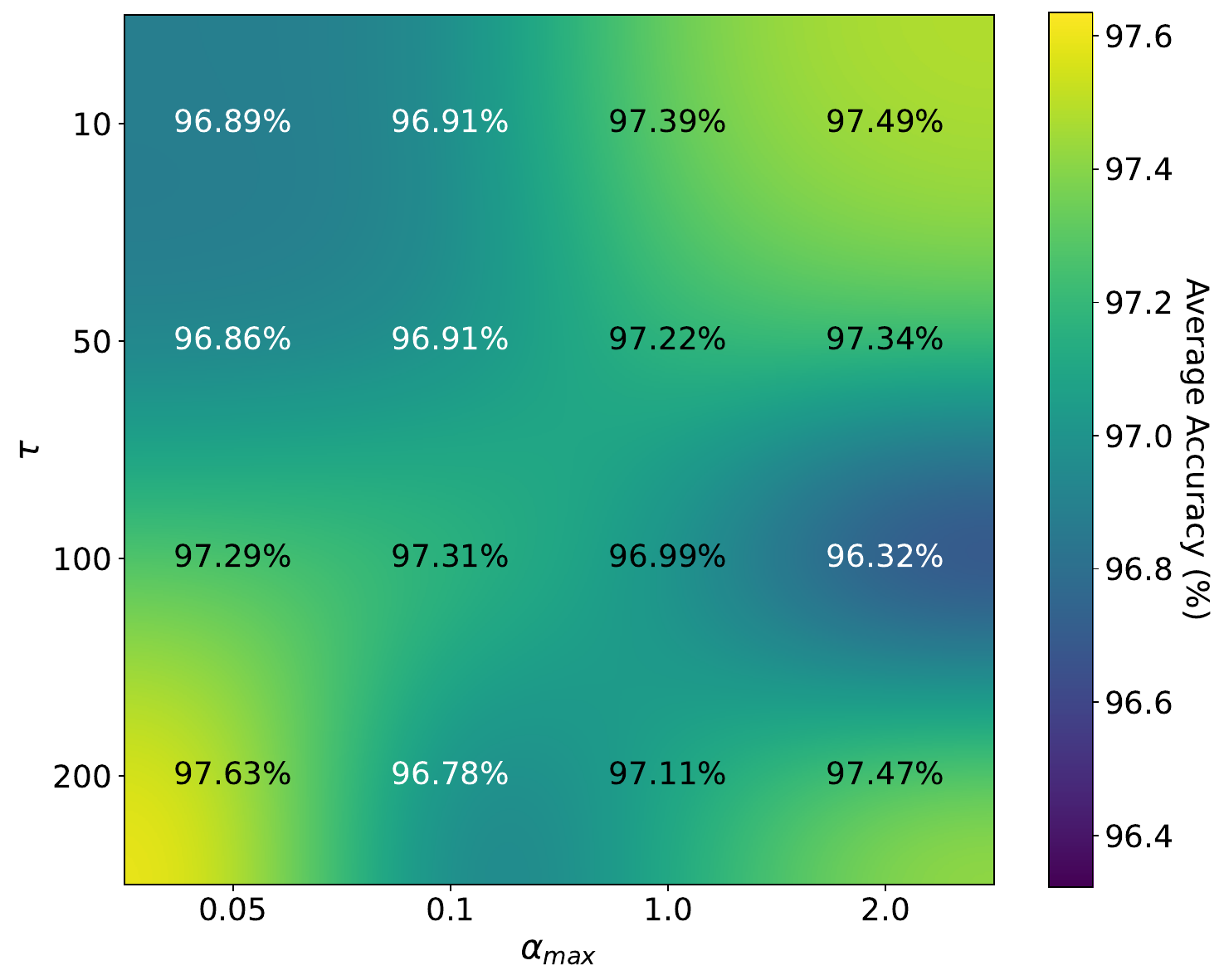}
    \caption{Heatmap of average accuracy for a FedSDAF variant \textbf{without MHSA}, using a dynamic distillation weight $\alpha_t$. The analysis explores different values for $\alpha_{\text{max}}$ and $\tau$ on the PACS dataset.}
    \label{fig:hyper_alpha}
\end{figure}

We performed a grid search for $\alpha_{\text{max}}$ and $\tau$ on the PACS dataset for this MHSA-free variant. The results are shown in Figure~\ref{fig:hyper_alpha}.

The analysis of this experiment leads to a clear conclusion. The heatmap reveals that the optimal performance for the MHSA-free variant is achieved with a high $\alpha_{\text{max}}$ (2.0) and a moderately fast rate of increase ($\tau=100$), yielding a peak average accuracy of approximately 97.08\%. While this result confirms that a carefully tuned dynamic distillation schedule can enhance performance, it crucially represents the performance ceiling for this ablated model configuration. Notably, this peak accuracy remains significantly below the performance of our full FedSDAF model, which incorporates the MHSA mechanism (as detailed in Table~\ref{tab:hard_domain_ablation}). This performance deficit underscores that the contribution of MHSA is not merely a matter of balancing feature fusion, which could be emulated by adjusting loss weights. Instead, it provides a distinct architectural advantage, likely by capturing complex inter-feature relationships that are essential for robust generalization. Therefore, this ablation study validates that while a dynamic distillation weight is a viable optimization strategy, it is not a substitute for the fundamental benefits conferred by the MHSA component, thereby justifying its integral role in our proposed framework.

\section{Conclusion}

In this paper, we introduced FedSDAF, a novel framework for federated domain generalization that challenges the conventional focus on domain invariance. Our work is motivated by the key insight, validated through a targeted study, that source-domain-aware features possess superior generalization power in data-isolated environments. FedSDAF operationalizes this insight through an innovative dual-adapter architecture, which decouples and synergizes client-specific "local expertise" and a shared "global consensus." A bidirectional knowledge distillation mechanism enables a collaborative dialogue between these components, allowing the framework to systematically transform unique source domain knowledge into a powerful signal for generalization. Extensive experiments on four challenging benchmarks demonstrate that FedSDAF significantly outperforms existing methods, establishing a new state-of-the-art.

Despite its strong performance, our approach has limitations. The dual-adapter architecture and bidirectional distillation process introduce a moderate increase in computational overhead on the client side during local training compared to simpler fine-tuning methods. Furthermore, the performance of FedSDAF, like other PEFT-based approaches, is inherently tied to the quality of the pre-trained foundation model. Future research could explore more computationally efficient distillation techniques or investigate the applicability of the source-aware paradigm to other modalities and federated learning challenges.


\end{document}